\documentclass[11pt,a4paper]{article}
\usepackage[hyperref]{emnlp-ijcnlp-2019}
\usepackage{times}
\usepackage{latexsym}
\usepackage{array}
\usepackage{amsmath}
\usepackage{amsfonts}
\usepackage{graphicx}
\usepackage{float}
\usepackage{CJKutf8}
\usepackage{color}
\usepackage{multirow}

\newcolumntype{L}[1]{>{\raggedright\arraybackslash}p{#1}}
\newcolumntype{C}[1]{>{\centering\arraybackslash}p{#1}}
\newcolumntype{R}[1]{>{\raggedleft\arraybackslash}p{#1}}

\usepackage{url}
\graphicspath{ {./imgs/} }

\aclfinalcopy % Uncomment this line for the final submission

\title{Generating Classical Chinese Poems from Vernacular Chinese }
\author{
Zhichao Yang$^{1}\thanks{\ \ Equal contribution}$, \ Pengshan Cai$^{1*}$, \ Yansong Feng$^2$, \ Fei Li$^1$, \\  \textbf{Weijiang Feng}$^3$, \textbf{Elena Suet-Ying Chiu}$^1$, \ \textbf{Hong Yu}$^{1}$ \\
$^{1}$ University of Massachusetts, MA, USA\\
{\tt \{zhichaoyang, pengshancai\}@umass.edu } {\tt foxlf823@gmail.com }  \\
{\tt chiu@llc.umass.edu } {\tt hong\_yu@uml.edu } \\
$^{2}$ Institute of Computer Science and Technology, Peking University, China \\
 {\tt fengyansong@pku.edu.cn } \\
$^{3}$ College of Computer, National University of Defense Technology, China\\
 {\tt fengweijiang14@nudt.edu.cn } \\
}

\date{}

\begin{document}

\maketitle

\begin{abstract}
Classical Chinese poetry is a jewel in the treasure house of Chinese culture. 
Previous poem generation models only allow users to employ keywords to interfere the meaning of generated poems, leaving the dominion of generation to the model. In this paper, we propose a novel task of generating classical Chinese poems from vernacular, which allows users to have more control over the semantic of generated poems. We adapt the approach of unsupervised machine translation (UMT) to our task. 
We use segmentation-based padding and reinforcement learning to address under-translation and over-translation respectively. 
According to experiments, our approach significantly improve the perplexity and BLEU compared with typical UMT models.
Furthermore, we explored guidelines on how to write the input vernacular to generate better poems.
Human evaluation showed our approach can generate high-quality poems which are comparable to amateur poems.

% Furthermore, we tried different inputs such as \emph{Song lyric ci} instead of vernacular and found the more similar the characteristics of input and classical poems are, the better the generated poems are.

% Based on the recently introduced unsupervised machine translation (UMT) framework, we propose phrase segmentation-based padding and a reinforcement learning based approach to  alleviate under-translation and over-translation respectively, allowing the model to preserve the intact semantic of the vernacular paragraph in the generated poem.
% Through extensive machine and human evaluation, we prove the effectiveness of our model and gain a few interesting insights from literature perspective.

%so that users may control the semantic of the poem through input text.

% while maintaining the meaning of the vernacular text, making poem generation semantic-controllable.

% Based on the unsupervised machine translation (UMT) framework, 
% we could bridge different forms of literature from across the history.
% We improve the  under-translation and over-translation issues, which impact the quality of generated poems. 
% We propose novel models to solve under-translation and over-translation, more severe problems when the source and target length differs compared to common neural machine translation.
% Experimental results show our models effectively reduce under-translation and over-translation, which in turn leads to better generated poems. 
\end{abstract}

\section{Introduction} % Cai Yang

% The classical Chinese poetry is a precious cultural heritage and it is a jewel in the treasure house of Chinese culture. During the history
% of more than two thousand years, millions of poems are written to describe  beautiful pastoral landscape, heroic spirit of defending the homeland, important religious festivals occasions, the chant for freedom of desire and pursuit of personal value, etc. Along with these topics, poetry has to follow some specific structural, tonal and rhythmical patterns in order to be considered as a masterpiece.

% Poetry generation is a popular field in Natural Language Processing. However, 
% generating traditional Chinese poems from modern Chinese sentences is never explored. Most of the state-of-art model could only generate poetry given some topic words. They generate poems by expanding topic words into sentences and sentences into poems. But as we know, the words used in ancient poems are different from modern languages. As a consequence, the existing methods may fail to generate meaningful poems given a modern term. However, when a long sentence is given as input, our model would interpret its meaning by not only the modern term but also the words surrounding them, and thus produce a better match with user's intention.

% yiyi
During  thousands of years, millions of classical Chinese poems have been written. They contain ancient poets' emotions such as their appreciation for nature, desiring for freedom and concerns for their countries.
% Amongst many types of Chinese poetry, \emph{quatrain poems} stand out. 
Among various types of classical poetry, \emph{quatrain poems} stand out.
% \fei{I can't understand why using `stand out` here. Maybe `\emph{quatrain poems} is the most classical.'}
On the one hand, their aestheticism and terseness exhibit unique elegance. 
On the other hand, composing such poems is extremely challenging due to their phonological, tonal and structural restrictions. 

% others limitation
% With the fast development of artificial intelligence, deep neural models are applied in classical Chinese poem composition. 
% Compared with human poets, neural models could process large amount of data and extract rich text patterns.

\begin{figure*}[h]
	\centering
	\includegraphics[width=1.05\textwidth]{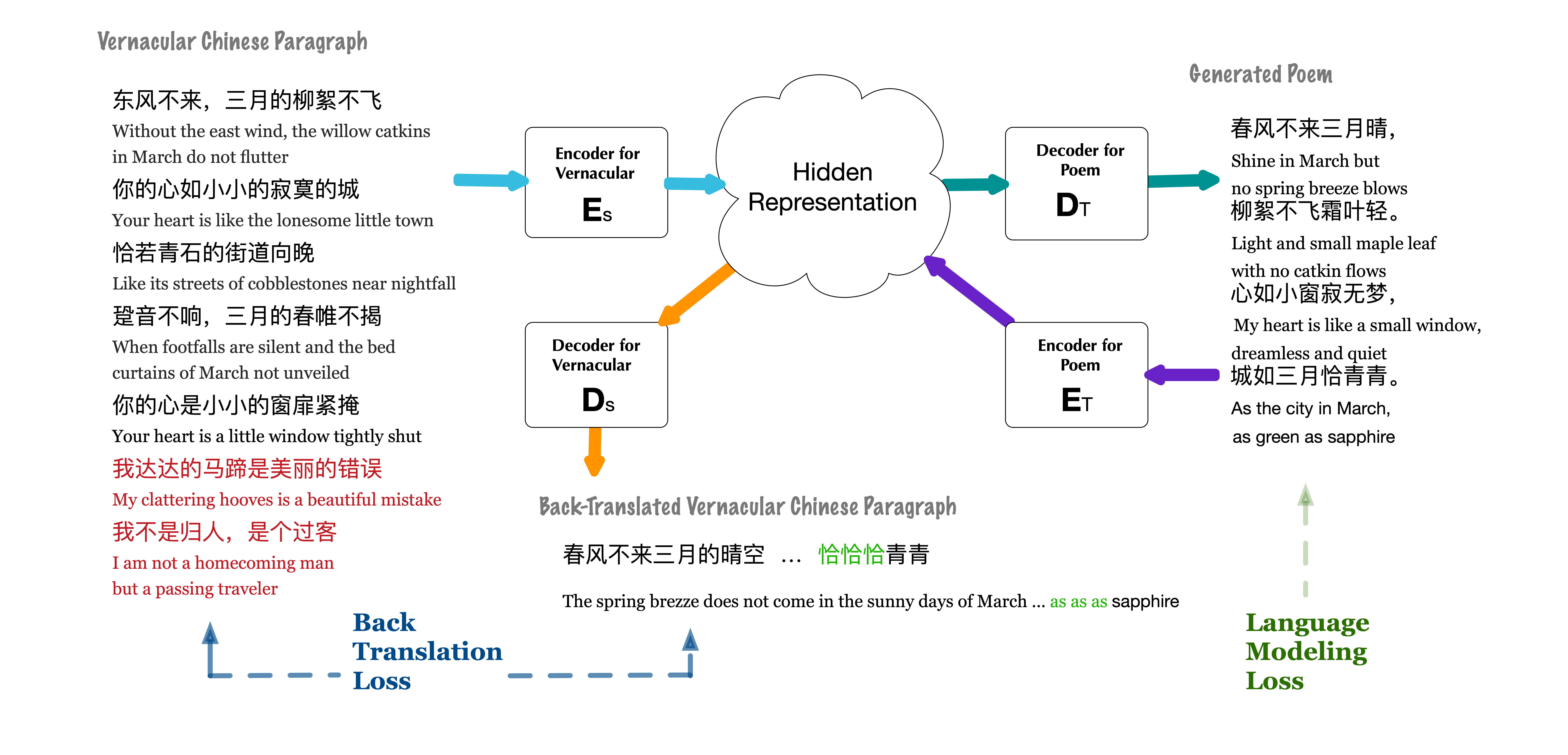}
	\caption{
	An example of the training procedures of our model. Here we depict two procedures, namely back translation and language modeling. Back translation has two paths, namely $\textbf{E}_S \rightarrow \textbf{D}_T \rightarrow \textbf{E}_T \rightarrow \textbf{D}_S$ and $\textbf{D}_T \rightarrow \textbf{E}_S \rightarrow \textbf{D}_S \rightarrow \textbf{E}_T$. Language modeling also has two paths, namely $\textbf{E}_T \rightarrow \textbf{D}_T$ and $\textbf{E}_S \rightarrow \textbf{D}_S$. Figure \ref{fig:frame} shows only the former one for each training procedure.
	}
	\label{fig:frame}
\end{figure*}

% we are awesome
Most previous models for generating classical Chinese poems \cite{He2012GeneratingCC, Zhang2014ChinesePG} are based on limited keywords or characters at fixed positions (e.g., acrostic poems). Since users could only interfere with the semantic of generated poems using a few input words, models control the procedure of poem generation. In this paper, we proposed a novel model for classical Chinese poem generation. As illustrated in Figure \ref{fig:frame}, our model generates a classical Chinese poem based on a vernacular Chinese paragraph.
% What do we expect the generated poems to be like ?
Our objective is not only to make the model generate aesthetic and terse poems, but also keep rich semantic 
%\fei{semantic is a adjective. I am not sure it can be used as noun.} 
of the original vernacular paragraph. Therefore, our model gives users more control power over the semantic of generated poems by carefully writing the vernacular paragraph. 
% Below is not a very good point
% Furthermore, our model bridges the form of classical poems with the content of contemporary literature, poems generated from fragments of contemporary literature not only demonstrates the beauty of classical poems, but also keeps the semantic of the original fragments. 

Although a great number of classical poems and vernacular paragraphs are easily available, there  exist only limited human-annotated pairs of poems and their corresponding vernacular translations. Thus, it is unlikely to train such poem generation model using supervised approaches. Inspired by unsupervised machine translation (UMT) \cite{Lample2018PhraseBasedN}, we treated our task as a translation problem, namely translating vernacular paragraphs to classical poems. 
% where we see poem generation as translating vernacular texts to classical poems.
%vernacular and poem as source and target language respectively. 

However, our work is not just a straight-forward application of UMT. In a training example for UMT, the length difference of source and target languages are usually not large, but this is not true in our task. Classical poems tend to be more concise and abstract, while vernacular text tends to be detailed and lengthy. Based on our observation on gold-standard annotations, vernacular paragraphs usually contain more than twice as many Chinese characters as their corresponding classical poems. Therefore, such discrepancy leads to two main problems during our preliminary experiments: (1) \textbf{Under-translation}: when summarizing vernacular paragraphs to poems, some vernacular sentences are not translated and ignored by our model. Take the last two vernacular sentences in Figure \ref{fig:frame} 
%(\fei{highlight the sentences in red or green. Current color is hard to be distinguished}) in Figure \ref{fig:frame} 
as examples, they are not covered in the generated poem. (2) \textbf{Over-translation}: when expanding poems to vernacular paragraphs, certain words are unnecessarily translated for multiple times. For example, the last sentence in the generated poem of Figure \ref{fig:frame}, \emph{as green as sapphire}, is back-translated as \emph{as green as as as sapphire}.

Inspired by the phrase segmentation schema in classical poems \cite{jia:1984}, we proposed the method of phrase-segmentation-based padding to handle with under-translation. By padding poems based on the phrase segmentation custom of classical poems, our model better aligns poems with their corresponding vernacular paragraphs and meanwhile lowers the risk of under-translation.
% padding-based technique to suppress under-translation, and a reinforcement learning based loss function to suppress over-translation. BLABLABLA......
Inspired by \newcite{Paulus2018ADR}, we designed a reinforcement learning policy to penalize the model if it generates vernacular paragraphs with too many repeated words. Experiments show our method can effectively decrease the possibility of over-translation.

The contributions of our work are threefold:

(1) We proposed a novel task for unsupervised Chinese poem generation from vernacular text.

(2) We proposed using phrase-segmentation-based padding and reinforcement learning to address two important problems in this task, namely under-translation and over-translation.

(3) Through extensive experiments, we proved the effectiveness of our models and explored how to write the input vernacular to inspire better poems. Human evaluation shows our models are able to generate high quality poems, which are comparable to amateur poems.
 
% \fei{Could our dataset and code be publicly available? That would be a great contribution. Or at least one of them is available?}

% 2. When applying UMT framework, we observe severe under-translation and over-translation issues when the levels of abstraction differs in source and target language. We further propose improved models to deal with each issue. 
 
% Our models could be further applied to UMT based models where abstraction level between source and target language differs.

% Contribution

\section{Related Works} % Cai

% Chinese poem generation, compare task
\textbf{Classical Chinese Poem Generation}
% Automatic classical Chinese poem generation has been a popular topic in the past decade. 
Most previous works in classical Chinese poem generation focus on improving the semantic coherence of generated poems. 
% As a statistical machine translation model, \cite{He2012GeneratingCC} uses mutual information to select next sentence with better consistency with previous sentences.
% Based on [A Summary of Rhyming Constraints of Chinese Poems], the generated poems were evaluated from four perspectives (Fluency, Rhyme, Coherence and Meaning); 
Based on LSTM, Zhang and Lapata \shortcite{Zhang2014ChinesePG} purposed generating poem lines incrementally by taking into account the history of what has been generated so far.
Yan \shortcite{Yan2016iPA} proposed a polishing generation schema, each poem line is generated incrementally and iteratively by refining each line one-by-one.
Wang et al. \shortcite{Wang2016ChinesePG} and Yi et al. \shortcite{Yi2018ChinesePG} proposed models to keep the generated poems coherent and semantically consistent with the user's intent. There are also researches that focus on other aspects of poem generation. (Yang et al. \shortcite{Yang2018StylisticCP} explored increasing the diversity of generated poems using an unsupervised approach. Xu et al. \shortcite{Xu2018HowII} explored generating Chinese poems from images. 
While most previous works generate poems based on topic words, our work targets at a novel task: generating poems from vernacular Chinese paragraphs. 

% Unsupervised MT as well as its derivatives, compare models
% Others do not face the challenge of UT and OT
\noindent\textbf{Unsupervised Machine Translation} Compared with supervised machine translation approaches \cite{Cho2014LearningPR, Bahdanau2015NeuralMT}, unsupervised machine translation \cite{Lample2018UnsupervisedMT, Lample2018PhraseBasedN} does not rely on human-labeled parallel corpora for training. This technique is proved to greatly improve the performance of low-resource languages translation systems. (e.g. English-Urdu translation). 
The unsupervised machine translation framework is also applied to various other tasks, e.g. image captioning \cite{Feng2018UnsupervisedIC}, text style transfer \cite{Zhang2018StyleTA}, speech to text translation \cite{Bansal2017TowardsST} and clinical text simplification \cite{Weng2019UnsupervisedCL}.
% making it possible for machine learning models to be trained for these tasks when limited human-labeled parallel data is available.
The UMT framework makes it possible to apply neural models to tasks where limited human labeled data is available.
However, in previous tasks that adopt the UMT framework, the abstraction levels of source and target language are the same. This is not the case for our task.

\noindent\textbf{Under-Translation \& Over-Translation}  Both are troublesome problems for neural sequence-to-sequence models. Most previous related researches adopt the coverage mechanism \cite{Tu2016ModelingCF, Mi2016CoverageEM, Sankaran2016TemporalAM}. However, as far as we know, there were no successful attempt applying coverage mechanism to transformer-based models \cite{Vaswani2017AttentionIA}.

%(Maybe to Introduction or Model) Compared to the above tasks, using the unsupervised machine translation framework for classical Chinese poem generation is faced with the following challenge: classical Chinese poems are famous for their high degree of conciseness and abstraction. Usually, a modern Chinese sentence translated from a 32-character quatrain would contain around 70 characters. 
%As a result, when translating modern Chinese texts to poems, the model should be summarizing, while when translating poems to modern texts, the model should be expanding. 
% The inequality of information in source language and target language would usually lead to over-translation (from  low-information sentences to rich-information sentence) and under-translation (the reverse way)

% However, we note none of the aforementioned tasks need to deal with the information inequality issue. If the unsupervised translation framework is to be applied to tasks including automatic memorization, natural language generation, etc., then the framework needs to overcome the gap of information inequality.

\section{Model}

\subsection{Main Architecture} 

% \textbf{Task Formalization} Formally, our poem generation task relies on a modern Chinese text corpus $T = \{T_1, T_2, ...T_{n_t}\}$ and a poem corpus $P = \{P_1, P_2, ...P_{n_P}\}$, 

% \textbf{General Architecture}

We transform our poem generation task as an unsupervised machine translation problem. As illustrated in Figure \ref{fig:frame}, based on the recently proposed UMT framework \cite{Lample2018PhraseBasedN}, our model is composed of the following components: 
\begin{itemize}
\setlength\itemsep{0em}
\item Encoder $\textbf{E}_s$ and decoder $\textbf{D}_s$ for vernacular paragraph processing
\item Encoder $\textbf{E}_t$ and decoder $\textbf{D}_t$ for classical poem processing
\end{itemize}
where $\textbf{E}_s$ (or $\textbf{E}_t$) takes in a vernacular paragraph (or a classical poem) and converts it into a hidden representation, and $\textbf{D}_s$ (or $\textbf{D}_t$) takes in the hidden representation and converts it into a vernacular paragraph (or a poem). Our model relies on a vernacular texts corpus $\textbf{\emph{S}}$
% $\textbf{\emph{S}} = \{S_1, S_2, ...S_{n_s}\}$ 
and a poem corpus $\textbf{\emph{T}}$.
% $\textbf{\emph{T}} = \{T_1, T_2, ...T_{n_t}\}$,
We denote $S$ and $T$ as instances in $\textbf{\emph{S}}$ and $\textbf{\emph{T}}$ respectively.

The training of our model relies on three procedures, namely \emph{parameter initialization}, \emph{language modeling} and \emph{back-translation}. We will give detailed introduction to each procedure.
%where there are two models: $M_{T2P}$ takes in a modern Chinese paragraph $T_i$ and outputs a poem $P_i$, $M_{P2T}$ takes in a poem $P_j$ and outputs a modern Chinese paragraph $T_j$. 

\noindent\textbf{Parameter initialization} 
% Words of two different languages that express similar meanings should also have similar neighboring structures, thus parameter initialization should help associates words with their plausible translations in the other language.  
As both vernacular and classical poem use Chinese characters, we initialize the character embedding of both languages in one common space, the same character in two languages shares the same embedding. This initialization helps associate characters with their plausible translations in the other language. 
% \fei{This paragraph is too short to be an independent section. Please give more details and using some equations is better.} 

\noindent\textbf{Language modeling} It helps the model generate texts that conform to a certain language. A well-trained language model is able to detect and correct minor lexical and syntactic errors. We train the language models for both vernacular and classical poem by minimizing the following loss: 

\begin{equation}
\begin{aligned}
\mathcal{L}^{lm} =& \mathop{{}\mathbb{E}}_{S \in \textbf{\emph{S}}} [ -  \log P(S|\textbf{D}_s(\textbf{E}_s(S_N))] + \\
                  & \mathop{{}\mathbb{E}}_{T \in \textbf{\emph{T}}} [ -  \log P(T|\textbf{D}_t(\textbf{E}_t(T_N))],
\end{aligned}
\end{equation}
where $S_N$ (or $T_N$) is generated by adding noise (drop, swap or blank a few words) in $S$ (or $T$).

\noindent\textbf{Back-translation} Based on a vernacular paragraph $S$, we generate a poem $T_S$ using $\textbf{E}_s$ and $\textbf{D}_t$, we then translate $T_S$ back into a vernacular paragraph $S_{T_S} = \textbf{D}_s(\textbf{E}_t(T_S))$. Here,  $S$ could be used as gold standard for the back-translated paragraph $S_{T_s}$. In this way, we could turn the unsupervised translation into a supervised task by maximizing the similarity between $S$ and $S_{T_S}$. The same also applies to using poem $T$ as gold standard for its corresponding back-translation $T_{S_T}$.
We define the following loss:

\begin{equation}
\begin{aligned}
\mathcal{L}^{bt} =& \mathop{{}\mathbb{E}}_{S \in \textbf{\emph{S}}} [ -  \log P(S|\textbf{D}_s(\textbf{E}_t(T_S))] + \\
                  & \mathop{{}\mathbb{E}}_{T \in \textbf{\emph{T}}} [ -  \log P(T|\textbf{D}_t(\textbf{E}_s(S_T))].
\end{aligned}
\end{equation}

Note that $\mathcal{L}^{bt}$ does not back propagate through the generation of $T_S$ and $S_T$ as we observe no improvement in doing so. When training the model, we minimize the composite loss:

\begin{equation}
\begin{aligned}
\mathcal{L} = \alpha_1 \mathcal{L}^{lm} +  \alpha_2 \mathcal{L}^{bt},
\end{aligned}
\end{equation}
where $\alpha_1$ and $\alpha_2$ are scaling factors.

\subsection{Addressing Under-Translation and Over-Translation}
During our early experiments, we realize that the naive UMT framework is not readily applied to our task. Classical Chinese poems are featured for its terseness and abstractness. They usually focus on depicting broad poetic images rather than details.
% thus they are usually much shorter than their corresponding vernacular translations. 
We collected a dataset of classical Chinese poems and their corresponding vernacular translations, the average length of the poems is $32.0$ characters, while for vernacular translations, it is $73.3$. The huge gap in sequence length between source and target language would induce over-translation and under-translation when training UMT models. In the following sections, we explain the two problems and introduce our improvements.

\subsubsection{Under-Translation}
% \fei{I think using alleviating decreases the contribution of your method.}
% zhichao sts

By nature, classical poems are more concise and abstract while vernaculars are more detailed and lengthy, to express the same meaning, a vernacular paragraph usually contains more characters than a classical poem. 
As a result, when summarizing a vernacular paragraph $S$ to a poem $T_S$, $T_S$ may not cover all information in $S$ due to its length limit.
In real practice, we notice the generated poems usually only cover the information in the front part of the vernacular paragraph, while the latter part is unmentioned.

\begin{figure}[h]
	\centering
	\includegraphics[width=0.51\textwidth]{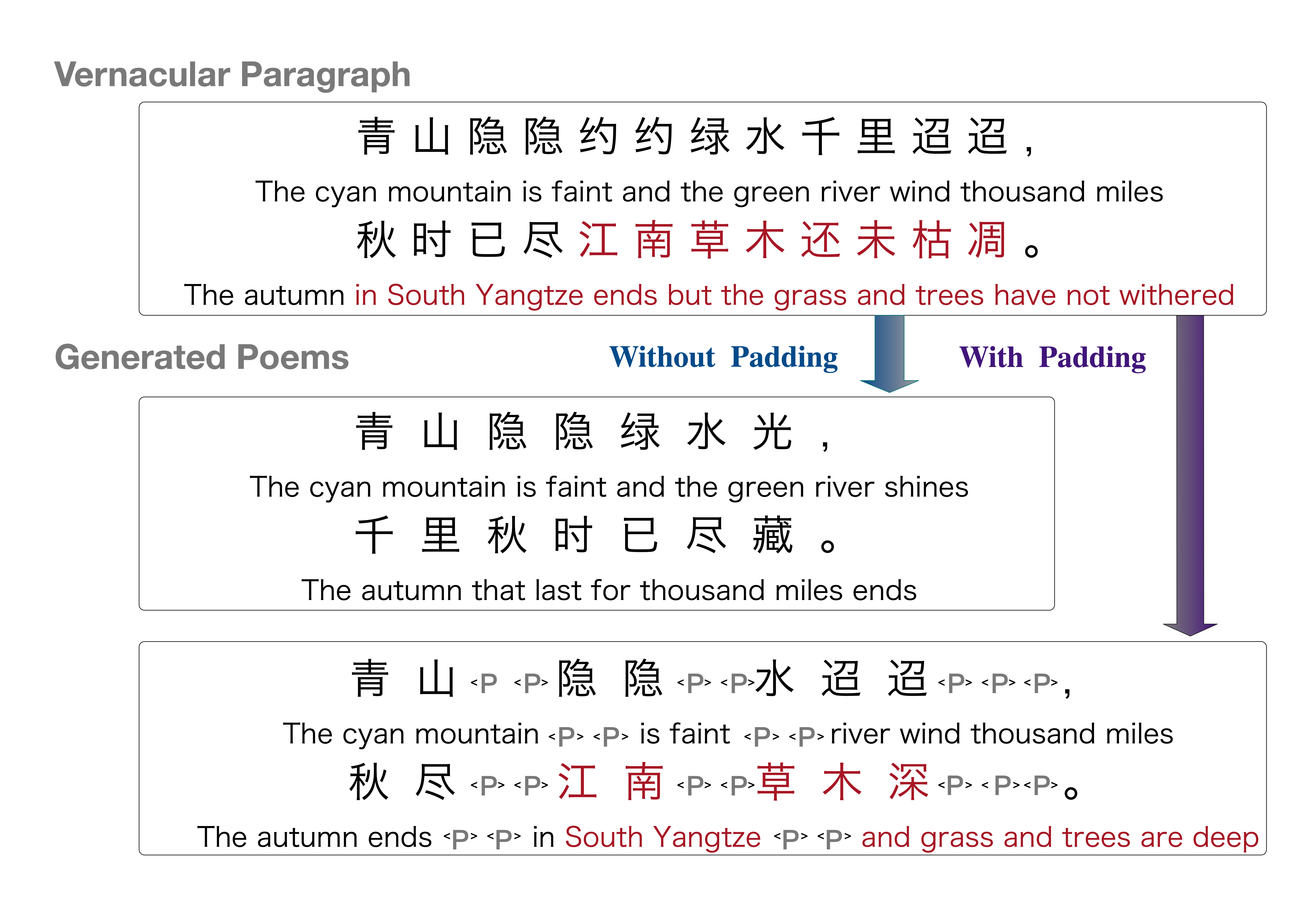}
	\caption{A real example to show the effectiveness of our phrase-segmentation-based padding. Without padding, the vernacular paragraph could not be aligned well with the poem. Therefore, the text \emph{in South Yangtze ends but the grass and trees have not withered} in red is not covered in the poem. By contrast, they are covered well after using our padding method. }
	\label{fig:padding_process}
\end{figure}

% To better align the vernacular paragraph with the poem, 
To alleviate under-translation, we propose phrase segmentation-based padding. 
Specifically, we first segment  each line in a classical poem into several sub-sequences, we then join these sub-sequences with the special padding tokens \textless p\textgreater. During training, the padded lines are used instead of the original poem lines. 
As illustrated in Figure \ref{fig:padding_process}, padding would create better alignments between a vernacular paragraph and a prolonged poem, making it more likely for the latter part of the vernacular paragraph to be covered in the poem.
As we mentioned before, the length of the vernacular translation is about twice the length of its corresponding classical poem, so we pad each segmented line to twice its original length.

According to Ye \shortcite{jia:1984}, to present a stronger sense of rhythm, each type of poem has its unique phrase segmentation schema, for example, most seven-character quatrain poems adopt the 2-2-3 schema, i.e. each quatrain line contains 3 phrases, the first, second and third phrase contains 2, 2, 3 characters respectively. Inspired by this law, we segment lines in a poem according to the corresponding phrase segmentation schema. In this way, we could avoid characters within the scope of a phrase to be cut apart, thus best preserve the semantic of each phrase.\cite{Chang2008OptimizingCW}
% (see Section 4.6.1 for details)

%  into phrases accordingly and pad the gap by their length. This process is [illustrated in figure ...]. Once the poem has been extended to almost the length of vernacular, we continue to train transformer baseline model with both Language Modeling and Back-translation steps mentioned above. 
% Because of the similar word style between both vernacular and poetry, the transformer would copy most phrases directly from vernacular to poem. Instead of simply copying every word in the first part and ran out of space, padding would help to copy from later part of vernacular. Even though this simply copied poem is hardly fluent, the poem model learned from later epochs would style it to look more like a poem. In general, these simply copied poem entities from padding would provide better translation coverage compared to those created from transformer baseline model, because the padding strategy extends the length of target poem to almost the length of modern vernacular.

\begin{table*} [h]
\begin{center}
\small
\begin{tabular}{ l C{2.7cm} C{2.7cm} C{2.7cm}} 
\hline
\textbf{} & Training set & Validation set & Test set \\
\hline
\# Poems & 163K & 19K & 487  \\
Average length of poems & 32.0 & 32.0 & 32.0 \\
\# vernacular paragraphs & 337K & 19K  & 487  \\
Average length of vernacular paragraphs & 71.8 & 76.8 & 73.3  \\
\hline
\end{tabular}
\caption{Statistics of our dataset}
\label{Tab:sta}
\end{center}
\end{table*}

\begin{table*} [h]
\begin{center}
\footnotesize
\begin{tabular}{ l | L{8.6cm} |L{5.3cm} } 
\hline
\textbf{ID} & \textbf{Vernacular paragraph} & \textbf{Generated poem} \\
\hline
1 & \begin{CJK*}{UTF8}{gkai}青山隐隐约约绿水千里迢迢，秋时已尽江南草木还未枯凋。二十四桥明月映照幽幽清夜，你这美人现在何处教人吹箫？\end{CJK*} & \begin{CJK*}{UTF8}{gkai}青山隐隐绿水光，千里秋时已尽藏。江南草木还未枯，二十四桥幽夜香。\end{CJK*}  \\
\hline
1 & Blue peak is blur and faint, green river meaders thousands miles,
the southern grass has not dried up in the autumn.
Twenty-four hole bridges and bright moons shine in the clear night.
where do you beautifully teach people to flute? & 
Blue peak blurs and green river shines,
thousands miles away, autumn has been hidden.
the southern grass has not dried up in the autumn.
Twenty-four hole bridges smells beautiful and quiet tonight.
 \\
\hline
\hline 
2 & \begin{CJK*}{UTF8}{gkai}拂袖起舞于梦中徘徊，相思蔓上心扉。她眷恋梨花泪，静画红妆等谁归，空留伊人徐徐憔悴。\end{CJK*} &
\begin{CJK*}{UTF8}{gkai}拂袖起舞梦徘徊，相思蔓上心扉开。玉眷梨花泪痕静，画红等谁归去来。\end{CJK*}  \\
\hline 
2 &
The sleeves danced in the dream, and the lovesickness was on the heart. 
She is in love with the tears of pears, 
and who is quietly wearing red makeup, 
only left alone to be languished slowly. &
The sleeves danced in the dream, 
the lovesickness appeared in the heart. 
Jade concerns tears of pears but the mark is still,
wearing red makeup waiting for the one to come and go. 
\\
\hline 
\hline 
3 & \begin{CJK*}{UTF8}{gkai}窗外的麻雀在电线杆上多嘴，你说这一句很有夏天的感觉。手中的铅笔在纸上来来回回，我用几行字形容你是我的谁。\end{CJK*} &
\begin{CJK*}{UTF8}{gkai}窗下麻姑灯火多，闲中说与万缘何。夏频手把铅华纸，来往回头用几多。\end{CJK*}  \\
\hline
3 &
The sparrow outside the window is talking on the pole. 
You say this sentence makes you feel very summer. 
The pencil in my hand is writing back and forth on the paper. 
I only use a few lines to describe who you are to me.&
Under the window lie sparrow girls in this prosperous city,
chit chatting about the destiny of the world.
Summer hands over many drawing canvas,
Looking back and forth, how many do you need? \\
\hline
\hline 
4 & \begin{CJK*}{UTF8}{gkai}雨天的屋瓦，浮漾湿湿的流光，灰而温柔，迎光则微明，背光则幽黯，对于视觉，是一种低沉的安慰。\end{CJK*} &
\begin{CJK*}{UTF8}{gkai}雨余屋瓦浮漾湿，流光灰色暖相迎。光则微明背则色，幽人黯黯对风清。\end{CJK*}  \\
\hline 
4 &
The rainy days of the roof tiles are soaking wet and wet, gray and gentle,
Facing the light, it is slightly bright,
Against the light, it is pitch dark,
For the concept of vision, it is a deep comfort. &
The excess rain makes roof tiles rippling,
ambilight gray is warm and welcoming.
Light is slightly bright, against is pure color.
The person hides in dark but faces wind breeze. \\
\hline 
\hline 
5 & \begin{CJK*}{UTF8}{gkai}只要历史不阻断，时间不倒退，一切都会衰老。老就老了吧，安详地交给世界一副慈祥美。假饰天真是最残酷的自我糟践。\end{CJK*} &
\begin{CJK*}{UTF8}{gkai}只要诸公不阻时，不倒退食一尘埃。会衰老矣安分世，一副慈祥假此来。\end{CJK*}  \\
\hline
5 &
As long as history does not block, time does not go backwards, 
everything will age.
It is fine to get old, and handing it to the world with kindness. 
Faking innocence is the cruelest self-destruction. &
As long as people do not block time, 
it will not go backwards and absorbs into a dust.
People should stay chill and get old.
Faking innocence is not the way to go.\\
\hline
\end{tabular}
\caption{A few poems generated by our model from  their corresponding vernacular paragraphs.}
\label{Tab:exp}
\end{center}
\end{table*}

% zhichao end

\subsubsection{Over-Translation} 
% As we mentioned before, 
In NMT, when decoding is complete, the decoder would generate an  \textless EOS\textgreater  token, indicating it has reached the end of the output sequence. 
However, when expending a poem $T$ into a vernacular Chinese paragraph $S_T$, due to the conciseness nature of poems, after finishing translating every source character in $T$, the output sequence $S_T$ may still be much shorter than the expected length of a poem‘s vernacular translation. As a result, the decoder would believe it has not finished decoding. Instead of generating the  \textless EOS\textgreater  token, the decoder would continue to generate new output characters from previously translated source characters. This would cause the decoder to repetitively output a piece of text many times. 

To remedy this issue, in addition to minimizing the original loss function $\mathcal{L}$, we propose to minimize a specific discrete metric, which is made possible with reinforcement learning. 

We define \emph{repetition ratio} $RR(S)$ of a paragraph $S$ as:

\begin{equation}
\begin{aligned}
RR(S) = 1 - \frac{vocab(S)}{len(S)},
\end{aligned}
\end{equation}
where $vocab(S)$ refers to the number of distinctive characters in $S$, $len(S)$ refers the number of all characters in $S$. Obviously, if a generated sequence contains many repeated characters, it would have high \emph{repetition ratio}.
Following the self-critical policy gradient training \cite{Rennie2017SelfCriticalST}, we define the following loss function:

\begin{equation}\label{eq:rr}
\begin{aligned}
\mathcal{L}^{rl} =& \mathop{{}\mathbb{E}}_{S \in \textbf{\emph{S}}} [ (RR(S_{T_S}) - \tau)  \log P(S|\textbf{D}_s(\textbf{E}_t(T_S))],
\end{aligned}
\end{equation}
where $\tau$ is a manually set threshold. Intuitively, minimizing $\mathcal{L}^{rl}$ is equivalent to maximizing the conditional likelihood of the sequence $S$ given $S_{T_S}$ if its \emph{repetition ratio} is lower than the threshold $\tau$.
Following \cite{Wu2016GooglesNM}, we revise the composite loss as:

\begin{equation}
\begin{aligned}
\mathcal{L'} = \alpha_1 \mathcal{L}^{lm} +  \alpha_2 \mathcal{L}^{bt} + \alpha_3 \mathcal{L}^{rl},
\end{aligned}
\end{equation}
where $\alpha_1, \alpha_2, \alpha_3$ are scaling factors. 
% In early stages of training, we set $\alpha_3$ as $0.0$ to first quickly obtain an initial model.

% Problem description with specific examples

% Solution$

\section{Experiment}

The objectives of our experiment are to explore the following questions: (1) How much do our models improve the generated poems? (Section \ref{sec:reborn}) (2) What are characteristics of the input vernacular paragraph that lead to a good generated poem? (Section \ref{sec:interpoetry}) (3) What are weaknesses of generated poems compared to human poems? (Section \ref{sec:humaneval}) To this end, we built a dataset as described in Section \ref{sec:dataset}. Evaluation metrics and baselines are described in Section \ref{sec:eval} and \ref{sec:baseline}. For the implementation details of building the dataset and models, please refer to supplementary materials.\footnote{Our data and code is publicly available at https://github.com/whaleloops/interpoetry}

% We evaluated our models on the following tasks: 

%\textbf{Rebirth of quatrains} aims to generate quatrains from modern Chinese text translations; \textbf{Interpoetry} aims to generate quatrains from other forms of Chinese literatures including \emp{prose}, \emph{}

%\footnote{Our experiment data and source code are available at www.anonymous.com}

\subsection{Datasets} 
% table, statistics
\label{sec:dataset}

\textbf{Training and Validation Sets} We collected a corpus of poems and a corpus of vernacular literature from online resources. The poem corpus contains 163K quatrain poems from \emph{Tang Poems} and \emph{Song Poems}, the vernacular literature corpus contains 337K short paragraphs from 281 famous books, the corpus covers various literary forms including prose, fiction and essay. Note that our poem corpus and a  vernacular corpus are not aligned. We further split the two corpora into a training set and a validation set. 

% Anything else?
% Should we emphasize our contribution of collecting the book corpus?

\noindent\textbf{Test Set} From online resources, we collected 487 seven-character quatrain poems from \emph{Tang Poems} and \emph{Song Poems}, as well as their corresponding high quality vernacular translations. These poems could be used as gold standards for poems generated from their corresponding vernacular translations. Table \ref{Tab:sta} shows the statistics of our training, validation and test set. 
% How to reflect the authority of our poem 

% \begin{center}
% \begin{tabular}{ |l|c|c|c|c| } 
% \hline
%  & Poems & average characters per poem & MCL paragraphs & average characters per MCL characters\\
% \hline
% Training Set &  &  &  &  \\
% Validation Set &  &  &  &  \\
% Test Set &  &  &  &  \\
% \hline
% \end{tabular}
% \end{center}

% \subsection{Implementation Details} % Yang % Don't Write Anything Here
% e.g. key parameters, directions
% Don't Write Anything Here
% Don't Write Anything Here
% Don't Write Anything Here
% will be simplified using supplementary material to make space for other content

\begin{table*} [h]
\begin{center}
\small
\begin{tabular}{ l|C{1.5cm}|C{1.5cm}|C{1.5cm}|C{1.5cm}|C{1.5cm}|C{1.5cm} } 
\hline
\textbf{Model} & \textbf{Perplexity} & \textbf{BLEU} & \textbf{BLEU-1} & \textbf{BLEU-2} & \textbf{BLEU-3} & \textbf{BLEU-4} \\
\hline
LSTM & 118.27 & 3.81 & 39.16 & 6.93 & 1.58 & 0.49 \\
%LSTM & 154.58 & 0.43 & 25.11 & 1.61 & 0.08 & 0.0 \\
Transformer &105.79 & 5.50 & 40.92 & 8.02 & 2.46 & 1.11  \\
+Anti OT & 77.33 & 6.08 & 41.22 & 8.72 & 2.82 & 1.36\\
+Anti UT & 74.21 & 6.34 & 42.20 & \textbf{9.04} & \textbf{2.96} & 1.44 \\
+Anti OT\&UT & \textbf{65.58} & \textbf{6.57} & \textbf{42.53} & 8.98 & \textbf{2.96} & \textbf{1.46}\\
\hline
\end{tabular}
\caption{Perplexity and BLEU scores of generating poems from vernacular translations. Since perplexity and BLEU scores on the test set fluctuates from epoch to epoch, we report the mean perplexity and BLEU scores over 5 consecutive epochs after convergence.}
\label{Tab:pplbleu}
\end{center}
\end{table*}

\begin{table*} [!htb]
\begin{center}
\footnotesize
\begin{tabular}{ l|C{2.0cm}|C{2.0cm}|C{2.0cm}|C{2.0cm}|C{2.0cm} } 
\hline
 \textbf{Model} & \textbf{\footnotesize Fluency} & \textbf{\footnotesize Semantic coherence}& \textbf{\footnotesize Semantic preservability}& \textbf{\footnotesize Poeticness} & \textbf{\footnotesize Total} \\
\hline
Transformer & 2.63 & 2.54 & 2.12 & 2.46 & 9.75  \\
+Anti OT & 2.80 & 2.75 & 2.44 & 2.71 & 10.70 \\
+Anti UT & 2.82 & 2.82 & 2.86 & 2.85 & 11.35 \\
+Anti OT\&UT & \textbf{3.21} & \textbf{3.27} & \textbf{3.27} & \textbf{3.28} & \textbf{13.13} \\
\hline
\end{tabular}
\caption{Human evaluation results of generating poems from vernacular translations. We report the mean scores for each evaluation metric and total scores of four metrics. 
%(Limited by our resources, we only ask human evaluators to grade transformer-based models)
}
\label{Tab:hm1}
\end{center}
\end{table*}

\subsection{Evaluation Metrics}
\label{sec:eval}

\noindent \textbf{Perplexity} Perplexity reflects the probability a model generates a certain poem. Intuitively, a better model would yield higher probability (lower perplexity) on the gold poem.

\noindent \textbf{BLEU} As a standard evaluation metric for machine translation, BLEU \cite{Papineni2001BleuAM} measures the intersection of n-grams between the generated poem and the gold poem. A better generated poem usually achieves higher BLEU score, as it shares more n-gram with the gold poem.
% Specifically, we use BLEU-4 as it's reflects the similarity not only in single tokens but also in higher grams. 

\noindent \textbf{Human evaluation} While perplexity and BLEU are objective metrics that could be applied to large-volume test set, evaluating Chinese poems is after all a subjective task. We invited 30 human evaluators to join our human evaluation. The human evaluators were divided into two groups. The expert group contains 15 people who hold a bachelor degree in Chinese literature, and the amateur group contains 15 people who holds a bachelor degree in other fields. All 30 human evaluators are native Chinese speakers.

We ask evaluators to grade each generated poem from four perspectives: 1) \emph{Fluency}: Is the generated poem grammatically and rhythmically well formed, 2) \emph{Semantic coherence}: Is the generated poem itself semantic coherent and meaningful, 3) \emph{Semantic preservability}: Does the generated poem preserve the semantic of the modern Chinese translation, 4) \emph{Poeticness}: Does the generated poem display the characteristic of a poem and does the poem build good poetic image. The grading scale for each perspective is from 1 to 5.

\subsection{Baselines}
\label{sec:baseline}

We compare the performance of the following models: (1) \emph{LSTM} \cite{Hochreiter1997LongSM}; (2)\emph{Naive transformer} \cite{Vaswani2017AttentionIA}; (3)\emph{Transformer + Anti OT} (RL loss); (4)\emph{Transformer + Anti UT} (phrase segmentation-based padding); (5)\emph{Transformer + Anti OT\&UT}.
% \begin{itemize}
% \item \emph{LSTM} \cite{Hochreiter1997LongSM}
% \item \emph{Naive transformer} \cite{Vaswani2017AttentionIA}
% \item \emph{Transformer + Anti OT} (RL loss)
% \item \emph{Transformer + Anti UT} (phrase segmentation-based padding)
% \item \emph{Transformer + Anti OT\&UT}
% \end{itemize}
% \fei{you can give more descriptions for these baselines.}

\subsection{Reborn Poems: Generating Poems from Vernacular Translations} 
% Experiment explaination
\label{sec:reborn}

As illustrated in Table \ref{Tab:exp} (ID 1).  Given the vernacular translation of each gold poem in test set, we generate five poems using our models.  Intuitively, the more the generated poem resembles the gold poem, the better the model is. We report mean perplexity and BLEU scores in Table \ref{Tab:pplbleu} (Where +Anti OT refers to adding the reinforcement loss to mitigate over-fitting and +Anti UT refers to adding phrase segmentation-based padding to mitigate under-translation), human evaluation results in Table \ref{Tab:hm1}.\footnote{We did not use LSTM in human evaluation since its performance is worse as shown in Table \ref{Tab:pplbleu}.}

% \begin{table} 

According to experiment results, perplexity, BLEU scores and total scores in human evaluation are consistent with each other. We observe all BLEU scores are fairly low, we believe it is reasonable as there could be multiple ways to compose a poem given a vernacular paragraph. 
% LSTM could not encode long range dependency, it may just 
% transformer-based models could better catch long-term dependencies. 
Among transformer-based models, both +Anti OT and +Anti UT outperforms the naive transformer, while Anti OT\&UT shows the best performance, this demonstrates alleviating under-translation and over-translation both helps generate better poems. Specifically, +Anti UT shows bigger improvement than +Anti OT. 
According to human evaluation, among the four perspectives, our Anti OT\&UT brought most score improvement in \emph{Semantic preservability}, this proves our improvement on semantic preservability was most obvious to human evaluators.
All transformer-based models outperform LSTM.
Note that the average length of the vernacular translation is over 70 characters, comparing with transformer-based models, LSTM may only keep the information in the beginning and end of the vernacular.
% our our Anti OT\&UT preserve more semantic in the generated poem from vernacular paragraph. 
We anticipated some score inconsistency between expert group and amateur group. However, after analyzing human evaluation results, we did not observed big divergence between two groups.

% \begin{table*} 
% \footnotesize
% \begin{center}
% \begin{tabular}{ |C{3cm}|C{1cm}|C{1cm}|C{1cm}|C{1cm}|C{1cm}|C{1cm}|C{1cm}|C{1cm}| } 
% \hline
% \textbf{Group} & \multicolumn{4}{c}{\textbf{Experts}} & \multicolumn{4}{|c|}{\textbf{Amateurs}}   \\
% \hline
% \textbf{Model} & Base & +AO & +AU &+AO/T & Base & +AO & +AU & +AO/T \\
% \hline
% \textbf{Fluency} &  &  &  &  &  &  &  &\\
% \textbf{Semantic Coherence} &  &  &  &  &  &  &  &\\
% \textbf{Semantic Preservity} &  &  &  &  &  &  &  &\\
% \textbf{Poeticness} &  &  &  &  &  &  &  &\\
% \textbf{Total} &  &  &  &  &  &  &  &\\
% \hline
% \end{tabular}
% \caption{Human evaluation results for poems generated from modern Chinese translations}
% \label{Tab:hm1}
% \end{center}
% \end{table*}

% Evaluation methods

% Analysis

\begin{table*} 
\begin{center}
\footnotesize
\begin{tabular}{ l|C{2.0cm}|C{2.0cm}|C{2.0cm}|C{2.0cm}|C{2.0cm} } 
\hline
\textbf{Literature form}  & \textbf{\footnotesize Fluency} & \textbf{\footnotesize Semantic coherence}& \textbf{\footnotesize Semantic preservability}& \textbf{\footnotesize Poeticness} & \textbf{\footnotesize Total} \\
\hline
Prose & 2.52 & 2.30 & 2.30 & 2.32 & 9.44\\
Modern poem & 2.37 & 2.34 & 2.01 & 2.16 & 8.88 \\
Pop song lyric & 2.40 & 2.31 & 2.24 & 2.42 & 9.37 \\
Song lyric & \textbf{2.62} & \textbf{2.54} & \textbf{2.26} & \textbf{2.49} & \textbf{9.91} \\
\hline
\end{tabular}
\caption{Human evaluation results for generating poems from various literature forms. We show the results obtained from our best model (Transformer+Anti OT\&UT).}
\label{Tab:hm2}
\end{center}
\end{table*}

\subsection{Interpoetry: Generating Poems from Various Literature Forms} 
\label{sec:interpoetry}
% Experiment explaination
Chinese literature is not only featured for classical poems, but also various other literature forms. \emph{Song lyric}\begin{CJK*}{UTF8}{gbsn}(宋词)\end{CJK*}, or \emph{ci} also gained tremendous popularity in its palmy days, standing out in classical Chinese literature. \emph{Modern prose}, \emph{modern poems} and \emph{pop song lyrics} have won extensive praise among Chinese people in modern days. The goal of this experiment is to transfer texts of other literature forms into quatrain poems. We expect the generated poems to not only keep the semantic of the original text, but also demonstrate terseness, rhythm and other characteristics of ancient poems. 
% Evaluation methods
Specifically, we chose 20 famous fragments from four types of Chinese literature (5 fragments for each of modern prose, modern poems, pop song lyrics and Song lyrics). We try to  As no ground truth is available, we resorted to human evaluation with the same grading standard in Section \ref{sec:reborn}.

% Analysis
Comparing the scores of different literature forms, we observe Song lyric achieves higher scores than the other three forms of modern literature. It is not surprising as both Song lyric and quatrain poems are written in classical Chinese, while the other three literature forms are all in vernacular. 

% Comparing the scores of poems generated from different paragraphs of the same literature form, 
Comparing the scores within the same literature form, we observe the scores of poems generated from different paragraphs tends to vary. After carefully studying the generated poems as well as their scores, we have the following observation: 

1) In classical Chinese poems, poetic images \begin{CJK*}{UTF8}{gbsn}(意象)\end{CJK*} were widely used to express emotions and to build artistic conception. A certain poetic image usually has some fixed implications.
For example, \emph{autumn} is usually used to imply sadness and loneliness.
%, while \emph{mandarin duck} is  a symbol of romance. 
However, with the change of time, poetic images and their implications have also changed. According to our observation, if a vernacular paragraph contains more poetic images used in classical literature, its generated poem usually achieves higher score.
As illustrated in Table \ref{Tab:exp}, both paragraph 2 and 3 are generated from pop song lyrics, paragraph 2 uses many poetic images from classical literature (e.g. pear flowers, makeup), while paragraph 3 uses modern poetic images (e.g. sparrows on the utility pole). Obviously, compared with poem 2, sentences in poem 3 seems more confusing, as the poetic images in modern times may not fit well into the language model of classical poems.

2) We also observed that poems generated from descriptive paragraphs achieve higher scores than from logical or philosophical paragraphs. 
For example, in Table \ref{Tab:exp}, both paragraph 4 (more descriptive) and paragraph 5 (more philosophical) were selected from famous modern prose. However, compared with poem 4, poem 5 seems semantically more confusing.
We offer two explanations to the above phenomenon: \textbf{i}. Limited by the 28-character restriction, it is hard for quatrain poems to cover complex logical or philosophical explanation. \textbf{ii}. As vernacular paragraphs are more detailed and lengthy, some information in a vernacular paragraph may be lost when it is summarized into a classical poem. While losing some information may not change the general meaning of a descriptive paragraph, it could make a big difference in a logical or philosophical paragraph.

\subsection{Human Discrimination Test} 
\label{sec:humaneval}
% Experiment explaination
%In our experiment we found that some of the generated poems are comparable to ones written by amateur poets. In order to quantify our observation, as well as to understand the difference between machine generated poems and human written poems, we carried out the following experiment: 
% Evaluation methods
We manually select 25 generated poems from vernacular Chinese translations and pair each one with its corresponding human written poem. 
% \footnote{See supplementary material for details}. 
We then present the 25 pairs to human evaluators and ask them to differentiate which poem is generated by human poet.\footnote{We did not require the expert group's participation as many of them have known the gold poems already. Thus using their judgments would be unfair.} 
% We report the experiment result in Table \ref{Tab:turing}.

% Analysis

% \begin{table*}[!htb]
% \begin{center}
% \small
% \begin{tabular}{ c|C{1.5cm}|C{1.5cm}|C{1.5cm}|C{1.5cm}|C{1.5cm}|C{1.5cm} } 
% \hline
% \textbf{Padding schema} & \textbf{Perplexity} & \textbf{BLEU} & \textbf{BLEU-1} & \textbf{BLEU-2} & \textbf{BLEU-3} & \textbf{BLEU-4} \\
% \hline
%  2-2-3  & \textbf{74.21} & \textbf{6.34} & \textbf{42.2} & \textbf{9.04} & \textbf{2.96} & \textbf{1.44}  \\
%  2-3-2  & 83.12 & 5.49 & 41.47 & 8.08 & 2.38 & 1.15 \\
%  3-2-2  & 85.66 & 5.75 & 41.84 & 8.34 & 2.54 & 1.22 \\
% \hline
% \end{tabular}
% \caption{Perplexity and BLEU scores of different padding schemas.}
% \label{Tab:paddingpplbleu}
% \end{center}
% \end{table*}

As demonstrated in Table \ref{Tab:vs}, although the general meanings in human poems and generated poems seem to be the same, the wordings they employ are quite different. This explains the low BLEU scores in Section 4.3.
According to the test results in Table \ref{Tab:turing}, human evaluators only achieved 65.8\% in mean accuracy. This indicates the best generated poems are somewhat comparable to poems written by amateur poets.

We interviewed evaluators who achieved higher than 80\% accuracy on their differentiation strategies. Most interviewed evaluators state they realize 
% we summarize their strategies as follows: 
% 1) Ancient Chinese poets prefer implicit narration of events and emotions, the straight forward way of narration let out the identity of generated poems.
the sentences in a human written poem are usually well organized to highlight a theme or to build a poetic image, while the correlation between sentences in a generated poem does not seem strong. As demonstrated in Table \ref{Tab:vs}, the last two sentences in both human poems (marked as red) echo each other well, while the sentences in machine-generated poems seem more independent.
% \textbf{3)} Some machine generated poems contains rarely used characters.
This gives us hints on the weakness of generated poems: 
% 1) Poem composition 
While neural models may generate poems that resemble human poems lexically and syntactically, it's still hard for them to compete with human beings in building up good structures. 

% \begin{table}[t]
% \begin{center}
% \footnotesize
% \begin{tabular}{  l |C{5.37cm} } 
% % \hline \textbf{Vernacular paragraph} & \textbf{Generated poem} \\
% \hline
% \multirow{2}{3em}{Human} & \begin{CJK*}{UTF8}{gkai}黄沙碛里客行迷，四望云天直下低。\end{CJK*} \\
%  & \begin{CJK*}{UTF8}{gkai}{\color{red}为言地尽天还尽，行到安西更向西}。\end{CJK*} \\
% \hline
% \multirow{2}{3em}{Machine} & \begin{CJK*}{UTF8}{gkai}异乡客子黄沙迷，雁路迷寒云向低\end{CJK*} \\
%  & \begin{CJK*}{UTF8}{gkai}{\color{black}只道山川到此尽，安西还要更向西}。\end{CJK*} \\
% \hline
% \hline
% \multirow{2}{3em}{Human} & \begin{CJK*}{UTF8}{gkai}绝域从军计惘然，东南幽恨满词笺。{\color{red}一箫一剑平生意，负尽狂名十五年
% }。\end{CJK*} \\
% \hline
% \multirow{2}{3em}{machine} & \begin{CJK*}{UTF8}{gkai}从军疆场志难酬，令人怅望东南州。{\color{black}形容仗剑敌平戎，情怀注满赋雪愁}。\end{CJK*}\\
% \hline
% \end{tabular}
% \caption{Examples of generated poems and their corresponding gold poems used in human discrimination test.} 
% % \fei{`' what is the purpose of this sentence?}
% \label{Tab:vs}
% \end{center}
% \end{table}

\begin{table}[t]
\begin{center}
\footnotesize
\begin{tabular}{  c |C{5.3cm} } 
% \hline \textbf{Vernacular paragraph} & \textbf{Generated poem} \\
\hline
\multirow{6}{*}{Human} & \begin{CJK*}{UTF8}{gkai}黄沙碛里客行迷，四望云天直下低。\end{CJK*}  \\
& \scriptsize Within yellow sand moraine guest travels lost,\\
& \scriptsize looking around found sky and clouds low. \\
& \begin{CJK*}{UTF8}{gkai}{\color{red}为言地尽天还尽，行到安西更向西}。\end{CJK*}  \\
& \scriptsize It is said that earth and sky ends here, \\ 
& \scriptsize however I need to travel more west than anxi. \\ 
\hline
\multirow{6}{*}{Machine} & \begin{CJK*}{UTF8}{gkai}异乡客子黄沙迷，雁路迷寒云向低。\end{CJK*}  \\
& \scriptsize Guest in yellow sand gets lost, \\
& \scriptsize geese are lost because clouds gets low. \\
& \begin{CJK*}{UTF8}{gkai}只道山川到此尽，安西还要更向西。\end{CJK*} \\
& \scriptsize It is said that mountains end here, \\ 
& \scriptsize however anxi is even more west.   \\ 
\hline
\hline
\multirow{6}{*}{Human} & \begin{CJK*}{UTF8}{gkai}绝域从军计惘然，东南幽恨满词笺。\end{CJK*} \\
& \scriptsize It’s hard to pay for the ambition of the military field,\\
& \scriptsize the anxiety of situation in southeast is all over poems. \\
& \begin{CJK*}{UTF8}{gkai}{\color{red}一箫一剑平生意，负尽狂名十五年}。\end{CJK*}  \\
& \scriptsize A flute and sword is all I care about in my life, \\ 
& \scriptsize 15 years have failed the reputation of "madman". \\ 
\hline
\multirow{6}{*}{Machine} & \begin{CJK*}{UTF8}{gkai}从军疆场志难酬，令人怅望东南州。\end{CJK*}  \\
& \scriptsize It’s hard to fulfill my ambition on the military field, \\
& \scriptsize the situation in the southeast states are troublesome. \\
& \begin{CJK*}{UTF8}{gkai}形容仗剑敌平戎，情怀注满赋雪愁。\end{CJK*} \\
& \scriptsize I would like to use my sword to conquer my enemy, \\ 
& \scriptsize yet my feelings are full of worry like the snow. \\ 
\hline
\end{tabular}
\caption{Examples of generated poems and their corresponding gold poems used in human discrimination test.} 
% \fei{`' what is the purpose of this sentence?}
\label{Tab:vs}
\end{center}
\end{table}

\begin{table}[t]
\begin{center}
\small
\begin{tabular}{ C{3.9cm}|C{1.9cm}} 
\hline
\textbf{Accuracy}& \textbf{Value} \\
\hline
Min& 52.0 \\

Max& 84.0 \\

Mean& 65.8 \\
\hline
\end{tabular}
\caption{The performance of human discrimination test.}
\label{Tab:turing}
\end{center}
\end{table}

% Analysis

\section{Discussion} 
\begin{figure}[t]
	\centering
	\includegraphics[width=0.5\textwidth]{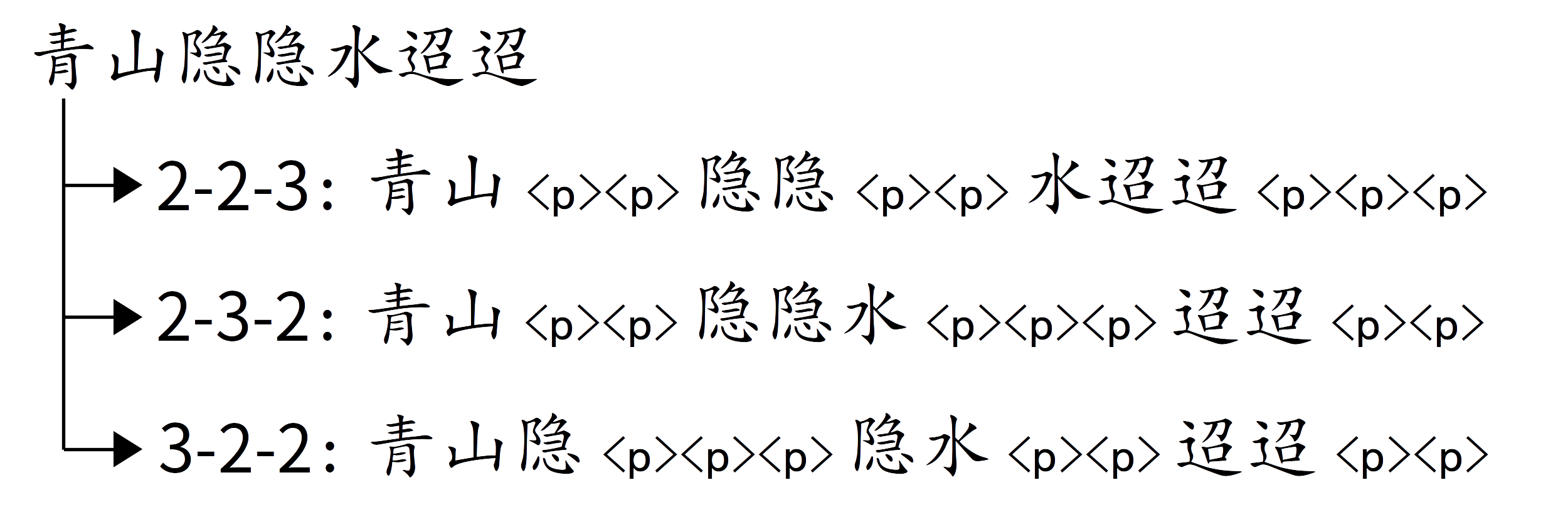}
	\caption{Examples of different padding schemas.}
	\label{fig:padding_detail}
\end{figure}

% Metre and rhymes
% zhichao sts
\textbf{Addressing Under-Translation} 
% There exists many ways to pad a sentence in poem to longer form. We now look in greater detail into the model’s performance with various padding schema. Rhythm based word segmentation in padding would enhance the integrity of semantic. As illustrated in examples below, an A-B-C padding schema means that a line in quatrain is split into 3 phrases, each with A, B, and C length long. According to Jia \cite{jia:1984}, most quatrains are separated in a 2-2-3 schema as this will present better rhythm aesthetic, which is one of many key evaluations of poem quality. In contrast to this phrase segmentation-based padding schema, we also pad in 2-3-2 and 3-2-2 patterns, which rarely exist in human written quatrains, to probe the importance of padding split position.
In this part, we wish to explore the effect of different phrase segmentation schemas on our phrase segmentation-based padding. 
% whether rhythm-based word segmentation in padding would enhance the integrity of semantic. 
According to Ye \shortcite{jia:1984},  most seven-character quatrain poems adopt the 2-2-3 segmentation schema.
% while other schemas are rarely used.
As shown in examples in Figure \ref{fig:padding_detail}, we compare our phrase segmentation-based padding (2-2-3 schema) to two less common schemas (i.e. 2-3-2 and 3-2-2 schema)
we report our experiment results in Table \ref{Tab:paddingpplbleu}.

The results show our 2-2-3 segmentation-schema greatly outperforms 2-3-2 and 3-2-2 schema in both perplexity and BLEU scores. Note that the BLEU scores of 2-3-2 and 3-2-2 schema remains almost the same as our naive baseline (Without padding). According to the observation, we have the following conclusions: 1) Although padding better aligns the vernacular paragraph to the poem, it may not improve the quality of the generated poem. 2) The padding tokens should be placed according to the phrase segmentation schema of the poem as it preserves the semantic within the scope of each phrase.

% Not all segmentation leads to significant higher BLEU score. We believe that phrase segmentation-based segmentation achieves better results by preserving semantic within the scope of each phrase.
% that careful word segmentation in padding would enhance the integrity of semantic.

\begin{table}[t]
\begin{center}
\small
\begin{tabular}{ c|C{1.5cm}|C{1.5cm}} 
\hline
\textbf{Padding schema} & \textbf{Perplexity} & \textbf{BLEU}  \\
\hline
 2-2-3  & \textbf{74.21} & \textbf{6.34}\\
 2-3-2  & 83.12 & 5.49  \\
 3-2-2  & 85.66 & 5.75  \\
\hline
\end{tabular}
\caption{Perplexity and BLEU scores of different padding schemas.}
\label{Tab:paddingpplbleu}
\end{center}
\end{table}

\noindent\textbf{Addressing Over-Translation} 
% Even though perplexity and BLEU score decrease when generating poem from vernacular for over translation model, would the RL method actually  alleviate the problem of repeatedly generating a previous part of vernacular from poem? 
% To answer this question, we utilize repetition ratio from equation [equation number]. The repetition ratio of original vernacular (written by human) is 30.1\% and RL method resulted in a 55\% decrease in repetition ratio on baseline transformer model (from 40.8\% to 34.9\%). This decrease is statistically significant, and show that less repetition is produced, suggesting that RL method  alleviate over translation problem.
To explore the effect of our reinforcement learning policy on alleviating over-translation, we calculate the \emph{repetition ratio} of vernacular paragraphs generated from classical poems in our validation set. We found \emph{naive transformer} achieves $40.8\%$ in repetition ratio, while our \emph{+Anti OT} achieves $34.9\%$. Given the repetition ratio of vernacular paragraphs (written by human beings) in our validation set is $30.1\%$, the experiment results demonstrated our RL loss effectively alleviate over-translation, which in turn leads to better generated poems.

% zhichao end

\section{Conclusion}
In this paper, we proposed a novel task of generating classical Chinese poems from vernacular paragraphs. We adapted the unsupervised machine translation model to our task and meanwhile proposed two novel approaches to address the under-translation and over-translation problems. Experiments show that our task can give users more controllability in generating poems. In addition, our approaches are very effective to solve the problems when the UMT model is directly used in this task. In the future, we plan to explore: (1) Applying the UMT model in the tasks where the abstraction levels of source and target languages are different (e.g., unsupervised automatic summarization); (2) Improving the quality of generated poems via better structure organization approaches.

% \noindent {\bf Preparing References:} \\

% Include your own bib file like this:
% {\small\verb|\bibliographystyle{acl_natbib}|
% \verb|\bibliography{emnlp-ijcnlp-2019}|}

% Where \verb|emnlp-ijcnlp-2019| corresponds to the {\tt emnlp-ijcnlp-2019.bib} file.
\clearpage
\bibliography{emnlp-ijcnlp-2019}

\begin{thebibliography}{26}
\expandafter\ifx\csname natexlab\endcsname\relax\def\natexlab#1{#1}\fi

\bibitem[{Bahdanau et~al.(2015)Bahdanau, Cho, and
  Bengio}]{Bahdanau2015NeuralMT}
Dzmitry Bahdanau, Kyunghyun Cho, and Yoshua Bengio. 2015.
\newblock Neural machine translation by jointly learning to align and
  translate.
\newblock In \emph{ICLR}.

\bibitem[{Bansal et~al.(2017)Bansal, Kamper, Lopez, and
  Goldwater}]{Bansal2017TowardsST}
Sameer Bansal, Herman Kamper, Adam Lopez, and Sharon Goldwater. 2017.
\newblock Towards speech-to-text translation without speech recognition.
\newblock In \emph{EACL}.

\bibitem[{Chang et~al.(2008)Chang, Galley, and Manning}]{Chang2008OptimizingCW}
Pi-Chuan Chang, Michel Galley, and Christopher~D. Manning. 2008.
\newblock Optimizing chinese word segmentation for machine translation
  performance.
\newblock In \emph{WMT@ACL}.

\bibitem[{Cho et~al.(2014)Cho, van Merrienboer, Çaglar G{\"u}lçehre,
  Bougares, Schwenk, and Bengio}]{Cho2014LearningPR}
Kyunghyun Cho, Bart van Merrienboer, Çaglar G{\"u}lçehre, Fethi Bougares,
  Holger Schwenk, and Yoshua Bengio. 2014.
\newblock Learning phrase representations using rnn encoder-decoder for
  statistical machine translation.
\newblock In \emph{EMNLP}.

\bibitem[{Feng et~al.(2019)Feng, Ma, Liu, and Luo}]{Feng2018UnsupervisedIC}
Yang Feng, Lin Ma, Wei Liu, and Jiebo Luo. 2019.
\newblock Unsupervised image captioning.
\newblock In \emph{The IEEE Conference on Computer Vision and Pattern
  Recognition (CVPR)}.

\bibitem[{He et~al.(2012)He, Zhou, and Jiang}]{He2012GeneratingCC}
Jing He, Ming Zhou, and Long Jiang. 2012.
\newblock Generating chinese classical poems with statistical machine
  translation models.
\newblock In \emph{AAAI}.

\bibitem[{Hochreiter and Schmidhuber(1997)}]{Hochreiter1997LongSM}
Sepp Hochreiter and J{\"u}rgen Schmidhuber. 1997.
\newblock Long short-term memory.
\newblock \emph{Neural Computation}, 9:1735--1780.

\bibitem[{Lample et~al.(2018{\natexlab{a}})Lample, Conneau, Denoyer, and
  Ranzato}]{Lample2018UnsupervisedMT}
Guillaume Lample, Alexis Conneau, Ludovic Denoyer, and Marc'Aurelio Ranzato.
  2018{\natexlab{a}}.
\newblock Unsupervised machine translation using monolingual corpora only.
\newblock In \emph{ICLR}.

\bibitem[{Lample et~al.(2018{\natexlab{b}})Lample, Ott, Conneau, Denoyer, and
  Ranzato}]{Lample2018PhraseBasedN}
Guillaume Lample, Myle Ott, Alexis Conneau, Ludovic Denoyer, and Marc'Aurelio
  Ranzato. 2018{\natexlab{b}}.
\newblock Phrase-based \& neural unsupervised machine translation.
\newblock In \emph{EMNLP}.

\bibitem[{Mi et~al.(2016)Mi, Sankaran, Wang, and
  Ittycheriah}]{Mi2016CoverageEM}
Haitao Mi, Baskaran Sankaran, Zhiguo Wang, and Abe Ittycheriah. 2016.
\newblock Coverage embedding models for neural machine translation.
\newblock In \emph{EMNLP}.

\bibitem[{Papineni et~al.(2001)Papineni, Roukos, Ward, and
  Zhu}]{Papineni2001BleuAM}
Kishore Papineni, Salim Roukos, Todd Ward, and Wei-Jing Zhu. 2001.
\newblock Bleu: a method for automatic evaluation of machine translation.
\newblock In \emph{ACL}.

\bibitem[{Paulus et~al.(2018)Paulus, Xiong, and Socher}]{Paulus2018ADR}
Romain Paulus, Caiming Xiong, and Richard Socher. 2018.
\newblock A deep reinforced model for abstractive summarization.
\newblock In \emph{ICLR}.

\bibitem[{Rennie et~al.(2017)Rennie, Marcheret, Mroueh, Ross, and
  Goel}]{Rennie2017SelfCriticalST}
Steven~J. Rennie, Etienne Marcheret, Youssef Mroueh, Jarret Ross, and Vaibhava
  Goel. 2017.
\newblock Self-critical sequence training for image captioning.
\newblock \emph{The IEEE Conference on Computer Vision and Pattern Recognition
  (CVPR)}, pages 1179--1195.

\bibitem[{Sankaran et~al.(2016)Sankaran, Mi, Al-Onaizan, and
  Ittycheriah}]{Sankaran2016TemporalAM}
Baskaran Sankaran, Haitao Mi, Yaser Al-Onaizan, and Abe Ittycheriah. 2016.
\newblock Temporal attention model for neural machine translation.
\newblock \emph{CoRR}, abs/1608.02927.

\bibitem[{Tu et~al.(2016)Tu, Lu, Liu, Liu, and Li}]{Tu2016ModelingCF}
Zhaopeng Tu, Zhengdong Lu, Yang~P. Liu, Xiaohua Liu, and Hang Li. 2016.
\newblock Modeling coverage for neural machine translation.
\newblock In \emph{ACL}.

\bibitem[{Vaswani et~al.(2017)Vaswani, Shazeer, Parmar, Uszkoreit, Jones,
  Gomez, Kaiser, and Polosukhin}]{Vaswani2017AttentionIA}
Ashish Vaswani, Noam Shazeer, Niki Parmar, Jakob Uszkoreit, Llion Jones,
  Aidan~N. Gomez, Lukasz Kaiser, and Illia Polosukhin. 2017.
\newblock Attention is all you need.
\newblock In \emph{NIPS}.

\bibitem[{Wang et~al.(2016)Wang, He, Wu, Wu, Li, Wang, and
  Chen}]{Wang2016ChinesePG}
Zhe Wang, Wei He, Hua Wu, Haiyang Wu, Wei Li, Haifeng Wang, and Enhong Chen.
  2016.
\newblock Chinese poetry generation with planning based neural network.
\newblock In \emph{COLING}.

\bibitem[{Weng et~al.(2019)Weng, Chung, and Szolovits}]{Weng2019UnsupervisedCL}
Wei-Hung Weng, Yu-An Chung, and Peter Szolovits. 2019.
\newblock \href {https://doi.org/10.1145/3292500.3330710} {Unsupervised
  clinical language translation}.
\newblock In \emph{Proceedings of the 25th ACM SIGKDD International Conference
  on Knowledge Discovery \&\#38; Data Mining}, KDD '19, pages 3121--3131, New
  York, NY, USA. ACM.

\bibitem[{Wu et~al.(2016)Wu, Schuster, Chen, Le, Norouzi, Macherey, Krikun,
  Cao, Gao, Klingner, Shah, Johnson, Liu, Kaiser, Gouws, Kato, Kudo, Kazawa,
  Stevens, Kurian, Patil, Wang, Young, Smith, Riesa, Rudnick, Vinyals, Corrado,
  Hughes, and Dean}]{Wu2016GooglesNM}
Yonghui Wu, Mike Schuster, Zhifeng Chen, Quoc~V. Le, Mohammad Norouzi, Wolfgang
  Macherey, Maxim Krikun, Yuan Cao, Qin Gao, Jeff Klingner, Apurva Shah, Melvin
  Johnson, Xiaobing Liu, Lukasz Kaiser, Stephan Gouws, Yoshikiyo Kato, Taku
  Kudo, Hideto Kazawa, Keith Stevens, George Kurian, Nishant Patil, Wei Wang,
  Cliff Young, Jason Smith, Jason Riesa, Alex Rudnick, Oriol Vinyals,
  Gregory~S. Corrado, Macduff Hughes, and Jeffrey Dean. 2016.
\newblock Google's neural machine translation system: Bridging the gap between
  human and machine translation.
\newblock \emph{CoRR}, abs/1609.08144.

\bibitem[{Xu et~al.(2018)Xu, Jiang, Qin, Wang, and Du}]{Xu2018HowII}
Linli Xu, Liang Jiang, Chuan Qin, Zhe Wang, and Dongfang Du. 2018.
\newblock How images inspire poems: Generating classical chinese poetry from
  images with memory networks.
\newblock In \emph{AAAI}.

\bibitem[{Yan(2016)}]{Yan2016iPA}
Rui Yan. 2016.
\newblock i, poet: Automatic poetry composition through recurrent neural
  networks with iterative polishing schema.
\newblock In \emph{IJCAI}.

\bibitem[{Yang et~al.(2018)Yang, Sun, Yi, and Li}]{Yang2018StylisticCP}
Cheng Yang, Maosong Sun, Xiaoyuan Yi, and Wenhao Li. 2018.
\newblock Stylistic chinese poetry generation via unsupervised style
  disentanglement.
\newblock In \emph{EMNLP}.

\bibitem[{Ye(1984)}]{jia:1984}
Jiaying Ye. 1984.
\newblock \emph{Poem Criticism with Jialin}.
\newblock Zhong hua shu ju, Beijing, China.

\bibitem[{Yi et~al.(2018)Yi, Sun, Li, and Yang}]{Yi2018ChinesePG}
Xiaoyuan Yi, Maosong Sun, Ruoyu Li, and Zonghan Yang. 2018.
\newblock Chinese poetry generation with a working memory model.
\newblock In \emph{IJCAI}.

\bibitem[{Zhang and Lapata(2014)}]{Zhang2014ChinesePG}
Xingxing Zhang and Mirella Lapata. 2014.
\newblock Chinese poetry generation with recurrent neural networks.
\newblock In \emph{EMNLP}.

\bibitem[{Zhang et~al.(2018)Zhang, Ren, Liu, Wang, Chen, Li, Zhou, and
  Chen}]{Zhang2018StyleTA}
Zhirui Zhang, Shuo Ren, Shujie Liu, Jianyong Wang, Peng Chen, Mu~Li, Ming Zhou,
  and Enhong Chen. 2018.
\newblock Style transfer as unsupervised machine translation.
\newblock \emph{CoRR}, abs/1808.07894.

\end{thebibliography}
\bibliographystyle{acl_natbib}

% \appendix

% \section{Supplemental Material}

\end{document}